\documentclass{article}
\usepackage{graphicx}
\usepackage{spconf,amsmath,graphicx}
\usepackage{wrapfig}
\usepackage{graphicx}
\usepackage{amsmath}

\usepackage{amssymb}
\usepackage{mathtools}
\usepackage{amsthm}
\usepackage{booktabs}
\usepackage[switch]{lineno}
\usepackage{xcolor}
\usepackage{colortbl}
\usepackage{multirow}
\usepackage[ruled]{algorithm2e}
\definecolor{bluex}{RGB}{54, 125, 189}
\usepackage[colorlinks=True,
            linkcolor=red,
            anchorcolor=blue,  
            pagebackref,  
            urlcolor=black,
            citecolor=bluex,
            ]{hyperref}

\title{Test-time Distribution Learning Adapter \\for Cross-modal Visual Reasoning}
\name{Yi Zhang$^1$\sthanks{Corresponding author.}, Ce Zhang$^2$}
\address{
\small{$^1$Feng AI Lab, Zhimeifenghua Smart Home, Guangzhou, China \quad  $^2$Carnegie Mellon University, Pittsburgh, United States}\\ 
\small{\texttt{yizhang.ai@ieee.org}\quad 
 \texttt{cezhang@cs.cmu.edu}}
}

\begin{document}
\ninept
\maketitle
%

\begin{abstract} 
Vision-Language Pre-Trained (VLP) models, such as CLIP, have demonstrated remarkable effectiveness in learning generic visual representations. Several approaches aim to efficiently adapt VLP models to downstream tasks with limited supervision, aiming to leverage the acquired knowledge from VLP models. However, these methods suffer from either introducing biased representations or requiring high computational complexity, which hinders their effectiveness in fine-tuning the CLIP model. Moreover, when a model is trained on data specific to a particular domain, its ability to generalize to uncharted domains diminishes. In this work, we propose \textit{\textbf{T}est-\textbf{T}ime \textbf{D}istribution Lear\textbf{N}ing \textbf{A}dapter (TT-DNA)} which directly works during the testing period. Specifically, we estimate Gaussian distributions to model visual features of the few-shot support images to capture the knowledge from the support set. The cosine similarity between query image and the feature distribution of support images is used as the prediction of visual adapter. Subsequently, the visual adapter's prediction merges with the original CLIP prediction via a residual connection, resulting in the final prediction. Our extensive experimental results on visual reasoning for human object interaction demonstrate that our proposed TT-DNA  outperforms existing state-of-the-art methods by large margins.
\end{abstract}
\begin{keywords}
Vision-Language Models, CLIP, Human Object Interaction, Few-Shot Learning, Visual Reasoning
\end{keywords}

\section{Introduction}
\label{sec:intro}
Recently, there has been a growing interest in large-scale vision-language pre-trained (VLP) models in both natural language processing and computer vision communities. These models leverage large-scale datasets comprising images and textual descriptions to learn joint representations of visual and textual information. Notably, VLP models, such as CLIP \cite{radford2021learning} and ALIGN \cite{jia2021scaling}, have shown remarkable performance on several visual tasks, including image recognition \cite{radford2021learning,gao2021clip}, object detection \cite{shi2022proposalclip,du2022learning}, and image captioning \cite{yao2021cpt}.

Fine-tuning is crucial for VLP models to adapt to various downstream tasks.
Although recent prompt tuning approaches, such as CoOp \cite{zhou2022learning}, ProDA \cite{lu2022prompt} and PLOT \cite{chen2023plot}, have demonstrated impressive performance on various benchmark tasks, these methods generally introduce a large number of parameters and require high computational complexity, making them time- and cost-consuming for practical implementation. To address this issue, Tip-Adapter \cite{zhang2022tip} has introduced a query-key cache model to obtain the adapter weights in a training-free manner, and the fine-tuned version, Tip-Adapter-F, has demonstrated remarkable performance on few-shot image classification. However, we recognize that Tip-Adapter only matches the test image feature against the feature of every single image in a class, which will result in introducing individual bias and hinder the generic representations of classes. Moreover, when a model is trained on data specific to a particular domain, its ability to generalize to uncharted domains diminishes.

\begin{figure}[t]
\centering
\includegraphics[width=\linewidth]{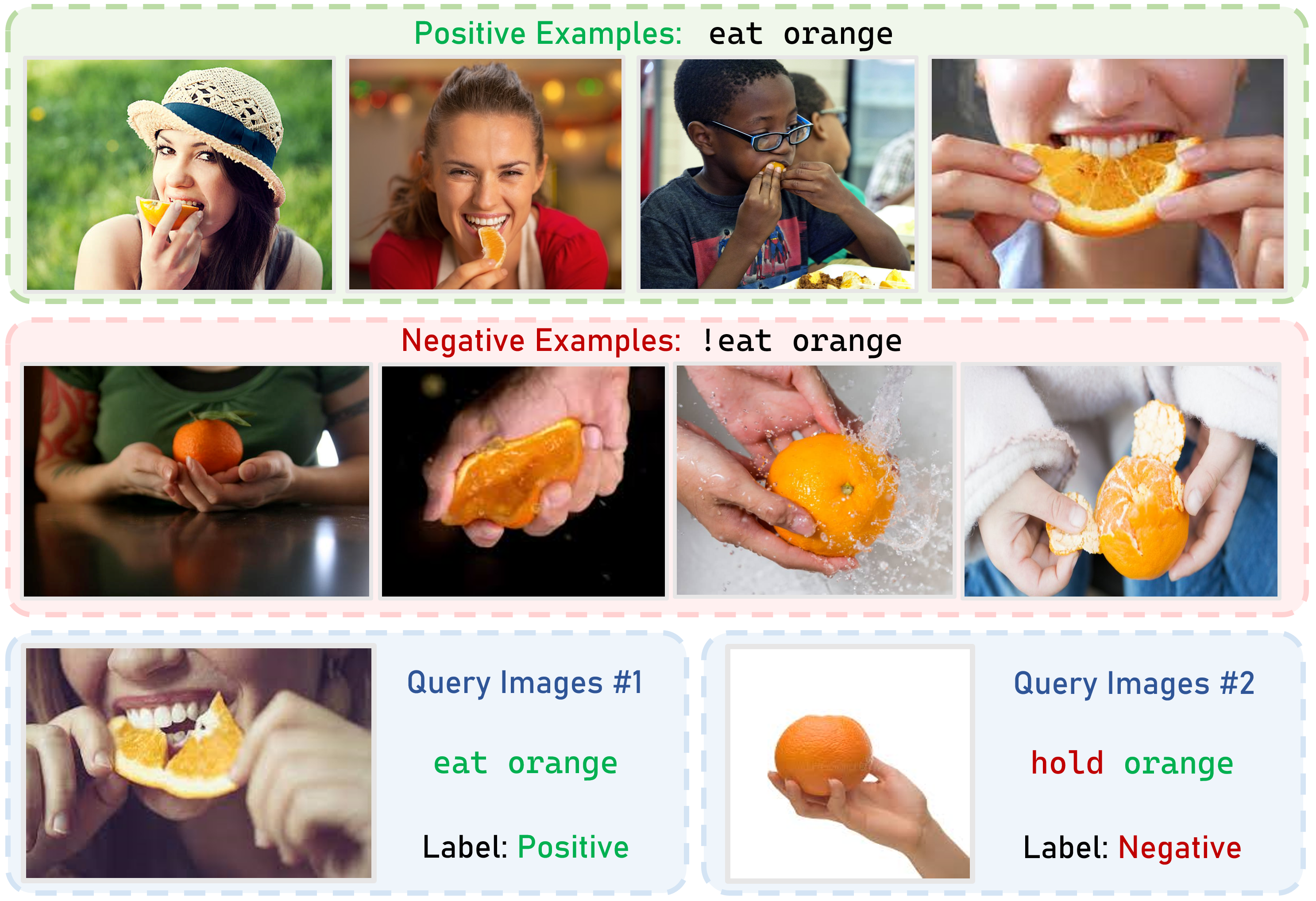}
\vspace{-20pt}
\caption{\textbf{Visual reasoning for human object interaction.} We present the task description of the Bongard-HOI \cite{jiang2022bongard} dataset. Please note that within the Bongard-HOI test set, there are a total of 6 positive examples, 6 negative examples, and a single query image. }
\label{fig:HOI}
\vspace{-13pt}
\end{figure}

To address these issues, in this work, we propose a method called Test-Time Distribution Learning Adapter (TT-DNA) which directly works during the testing period. Our TT-DNA method matches the query image feature against the feature distribution of support images instead of the feature of each individual image in a given class, which provides more comprehensive and generic class-specific representations for image recognition. 
Specifically, we estimate Gaussian distributions to model visual features of the few-shot support images to capture the knowledge from the support set. The cosine similarity between query image and the feature distribution of support images is used as the prediction of visual adapter. Subsequently, the visual adapter's prediction integrates with the original CLIP prediction through a residual connection. This enables the training-free TT-DNA to concurrently utilize knowledge from both the pre-trained CLIP and the few-shot support set. To demonstrate the effectiveness of our proposed DNA method, we apply and evaluate it on the challenging Bongard-HOI dataset, known as a visual reasoning benchmark emphasizing the compositional learning of human-object interactions (HOIs) from natural images. The training-free TT-DNA yields better performance than all the non-CLIP conventional methods and CLIP-based methods TPT~\cite{shu2022tpt} and BDC-Adapter~\cite{zhang2023bdc}, except 1 test scenario. To boost TT-DNA's performance,  we initialize a query model with the feature distributions of support images, which can be updated by back-propagation to provide better classification reference for the query image. We term this fine-tuned version as TT-DNA-F.
On top of that, our TT-DNA-F can be further enhanced by incorporating a textual adapter that tunes a set of additional parameters to learn optimal text feature. Our extensive experimental results demonstrate that our proposed TT-DNA-F outperforms existing state-of-the-art methods by large margins.
We provide a few-shot learning example from the Bongard-HOI benchmark in  Figure \ref{fig:HOI}.


\label{sec:method}
\begin{figure*}[!th]
\centering
\includegraphics[width=\textwidth]{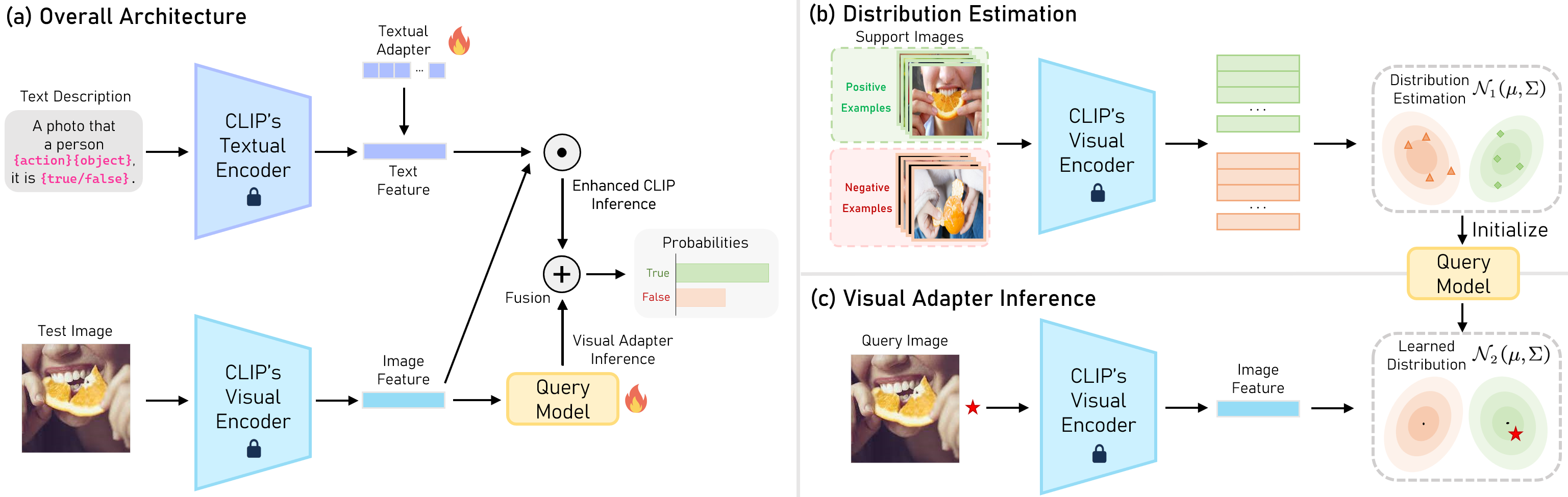}
\vspace{-18pt}
\caption{\textbf{Overview of the architecture of DNA}. The figure presents (a) the overall architecture of DNA, (b) the distribution estimation and (c) the visual adapter inference process. 
The fire icon signifies that the parameters will undergo updates during the training process.
}
\vspace{-8pt}
\label{fig:overview}
\end{figure*}

\section{Method}
\subsection{Background}
\textbf{Contrastive Language-Image Pre-training (CLIP).}
The CLIP model, comprising parallel encoders for image and text processing, employs a contrastive loss function during training. This function encourages similarity between image and text feature vectors, aligning both modalities within a joint embedding space. Represented as $\{E_t, E_v\}$, where $E_t$ is the text encoder and $E_v$ is the image encoder, CLIP is subsequently employed in zero-shot downstream tasks by utilizing a manually crafted prompt. Specifically focusing on image classification, this involves providing a single test image $X_{test}$ belonging to a specific class $y$, where $X_{test}\in \mathbb{R} ^{C\times H\times W}$ and $y \in \mathbb{R} ^ K$ for a $K$-class classification problem.
In the zero-shot baseline setting, each $y_i$ in the set $Y=\{y_1, y_2, \cdots, y_K\}$ is concatenated with a predefined prompt like $\rho =$ "a photo of," forming class-specific textual inputs $\{\rho; y_i\}$. Text features $\{t_1, t_2, \cdots, t_K\}$ are then generated using the text encoder $E_t$, where $t_i = E_t(\{\rho; y_i\})$. Following this, each text feature $t_i$ is combined with the image feature $v=E_v(X_{test})$ to compute a cosine similarity score. This score aids in predicting the probability of $X_{test}$ belonging to class $y_i$.
The prediction probability on $X$ can be denoted by 
\begin{equation}
\label{eq-clip}
   p(y_i|X_{test})=\frac{\exp \left( sim\left( t_i,v \right) /\tau \right)}{\sum\nolimits_{j=1}^K{\exp \left( sim\left( t_j,v \right) /\tau \right)}}, 
\end{equation}
where $\tau$ is the temperature of the softmax function.

\textbf{Visual Reasoning on Human Object Interaction.}
In the realm of context-dependent visual reasoning, as demonstrated in tasks like the Bongard-HOI task \cite{jiang2022bongard}, a test sample typically includes two sets of support images alongside a query image. These support image sets are designed to showcase the existence or non-existence of a human-object interaction (HOI) concept, for instance, something like "eat orange". Subsequently, the model is  tasked with the inference of whether the HOI concept is present in the query image. In this particular task, each concept is represented as a visual relationship, denoted as c = ⟨s, a, o⟩, where s denotes the subject (which is ``human" for HOI tasks), a denotes the action, and o represents the object involved. Every test sample, represented as $X_{test}$, encapsulates a specific concept by presenting c = ⟨s, a, o⟩ in one set of support images, which are considered positive examples. The other set of support images, on the other hand, serves as negative examples and demonstrates $\hat{c} = ⟨s, \hat{a}, o⟩$, where $\hat{a}$ is distinct from a. It is worth noting that neither the object o nor the action a is explicitly provided in the given task, and therefore, it is incumbent upon the model's reasoning ability to make predictions regarding the presence or absence of the featured concept c within the query image of the test sample. Prior research \cite{nie2020bongard, chen2020new} have tackled the Bongard-HOI problem by training the model on a diverse set of analogous tasks, utilizing the Bongard-HOI training split, thereby enabling it to make comparable inferences on test samples during test time. In the context of this task, the application of CLIP does not involve the utilization of supplementary training data, as the CLIP model has already acquired an extensive understanding of numerous visual concepts. As such, CLIP serves as a suitable choice for this form of visual reasoning task.

\subsection{Test-time Distribution Learning Adapter (TT-DNA)}
Figure \ref{fig:overview} shows an overview of our proposed TT-DNA method. In this section, we present our TT-DNA method in  detail. 

\textbf{Problem Definition.} Given a HOI test set which contains a query image $X_q$ and a support set which consists of $M$ positive images and $M$ negative images, the goal is to predict whether the query image belongs to the positive  or negative category. Obviously, it is a few-shot binary prediction problem.


\textbf{Estimating Visual Features’ Distribution.}
The visual features with the same class, which are generated from visual backbone $E_v$, are adjacent and can be modeled with the Gaussian distribution \cite{liu2021domain,lu2022prompt}. Also, recent works \cite{liu2022category,zhang2021adversarial,wang2021regularizing} 
demonstrate that the Gaussian distribution is effective in modeling the representations learned by neural networks. The mean and covariance of the distribution can be estimated from the generated visual features. We denote $V^P\triangleq \{v_{m}^{P}\}_{m=1}^{M}$ as the visual features of $M$ images within positive category, which are generated by $E_v$. Next, we can estimate 
the Gaussian distribution of positive category by $V_P$, denoted as $\mathcal{N} \left( \mu_P ,\Sigma_P \right)$.
Similarly, we estimate the Gaussian distribution of negative category, denoted as $\mathcal{N} \left( \mu_N ,\Sigma_N \right)$. We name them basic class distributions.





\textbf{Distribution Bias Diminishing.}  In the Distribution Estimation period, we obtain the basic class distributions. Nevertheless, in scenarios with limited data, the computed distributions tend to exhibit bias away from the desired expected distributions we aim to find. We define the distribution bias as,
\begin{equation}
    B_{dist} = {\mathbb{E}}_{X^{'} \sim D_{X^{'}}}[X^{'}] - {\mathbb{E}}_{X \sim D_X}[X]
\label{intra-bias-equation}
\end{equation}
where $D_{X^{'}}$ is the distribution of all data belonging to a specific class and $D_X$, signifying the distribution of the labeled data available for this class. It's evident that the expectations of these two distributions differ, with the disparity becoming more pronounced in low-data regimes. The anticipated distribution is ideally characterized by the mean and covariance of features from all class samples. However, in practical scenarios, only very limited samples are accessible as the support set, making it nearly impossible to capture the expected distribution fully. In the test set, we have only $M$ samples per support class, significantly fewer than the expected quantity. Consequently, due to this scarcity of samples, the class distribution tends to exhibit noticeable bias.

Drawing inspiration from Zhang \textit{et al.}~\cite{zhang2023prompt}, we leverage zero-shot DALL-E~\cite{Mini_Dalle} to create synthetic images, aiming to alleviate class distribution bias by enhancing the support set. Using a basic template like "a photo that person \{action\} \{object\}" for various categories, we generate synthetic images. Subsequently, we employ CLIP to filter the top-$K$ highest-quality images, expanding the support samples for each category. This process enables us to gather $J$-category ($M+K$)-sample training images, as expressed by the following formulation,
\begin{align}
    I_{J,(M+K)}= \{\text{DALL-E}(T_{J}),\ \ I_{J,M}\}, J \in \{ P, N\}
   \label{eq:dalle}
\end{align}
where $T_J$ denotes the $J$-category textual inputs, $I_{J,M}$ means that there are M images in  $J$ class, $P$ and $N$ are postive and negative respectively. In order to the few-shot setting, we set $K=M$.
Leveraging pre-trained language models for data expansion enables entirely zero-shot augmentation, eliminating the need for manual data collection or annotation. This approach inherently mitigates the challenges posed by data scarcity in few-shot learning.
The rectified distribution of a class is thus computed from the new support set, denoted as $\mathcal{N}_{s}^*\triangleq \{\mathcal{N} \left( \mu_J^* ,\Sigma_J^* \right)\},  J \in \{ P, N\}$. 
Compared with the basic class distributions $\mathcal{N}_{s}$, the rectified class distribution  $\mathcal{N}_{s}^*$ is closer to the expected distribution. 

\textbf{Visual Adapter Inference.}
During inference, the $L2$ normalized feature $f_v \in \mathbb{R}^{1 \times C}$ of the query image, which is first generated by visual encoder $E_v$. The affinities \cite{ru2022learning} between the query image and rectified distributions can be estimated as
\begin{equation}
\label{eq-H}
    H=\exp \left( -\eta \left(1-sim( f_v,\mathcal{N} _{s}^*)\right) \right),
\end{equation}
We utilize the exponential function to transform similarities into non-negative values, with $\eta$ controlling its degree of sharpness.
$sim\left( f_v,\mathcal{N} _{s}^* \right)$ represents the cosine similarity between test feature $f_v$ and rectified distributions $\mathcal{N} _{s}^*$. In our experiments, we find that simply using the means of distribution  to calculate the cosine similarity works well, we denote it as
\begin{equation}
\label{eq-sim-vmu}
    sim\left( f_v,\mathcal{N}_{s}^* \right)= f_v \hat{\mu^*},
\end{equation}
where $ \hat{\mu^*}\triangleq \{ \mu_P^*,\hat{\mu_N^*} \}$, which are the means of distributions $\mathcal{N}_{s}^*$. Since both $f_v$ and $\mathcal{\hat{\mu^*}}$ are $L2$-normalized, $f_v\hat{\mu^*}$ stands for their cosine similarity shown in Equation (\ref{eq-H}), so the output logits of the query image by visual adapter can be written as:
\begin{equation}
\label{eq-H-new}
    L_v=\exp \left( -\eta (1- f_v\hat{\mu^*}) \right).
\end{equation}

In brief, in the inference process of the visual adapter, we calculate the cosine similarity between the feature of the query image and the support features’ Gaussian distributions. Logits are subsequently generated according to the similarity scores between the feature of query image and the feature distributions of each class.
The query image is predicted as the class whose distribution has the highest similarity score with the query image.

\textbf{Inference Fusion.}
Finally, we fuse the inference from the visual adapter and original CLIP to better the prediction. We construct the input text like ``a photo that person \{action\}\{object\}, it is\{class\}", the class is \textbf{True} or \textbf{False}. Then we feed the text into text encoder $E_t$ and obtain the text feature $f_t$. 
Therefore, we combine the logits of the visual adapter and original CLIP, and the total logits of the query image $v$ by TT-DNA are calculated as 
\begin{align}
\label{eq-LOGIT}
\mathrm{Logits} &= \lambda L_v+L_C \nonumber \\ 
                &=\lambda \exp \left( -\eta \left( 1-f_v\hat{\mu^*} \right) \right) +f_vf_t^{\top}.
\end{align}
where $\lambda$ is the residual ratio and $f_t$ is the text features generated by $E_t$, the logits of pre-trained CLIP is calculated by $f_vf_t^{\top}$ . 

\subsection{TT-DNA with Fine Tuning (TT-DNA-F)}
\textbf{Query Model.}
While TT-DNA significantly enhances CLIP by integrating new knowledge from the support set, untrained TT-DNA falls behind training-dependent methods like TPT \cite{shu2022tpt} and BDC-Adapter \cite{zhang2023bdc} in a specific test scenario. To mitigate the gap while preserving the efficiency, we propose TT-DNA-F, which treats the means of class distributions as a good initialization of a query model, which can be denoted as,
\begin{equation}
    Q\left( x \right) =\exp \left( -\eta (1-Wx) \right).
\end{equation}
We initialize the model with $ \hat{\mu^*}\triangleq \{ \mu_P^*,\hat{\mu_N^*} \}$ as defined in Equation (\ref{eq-H-new}). The idea is that enhancing the means of class distributions can improve affinity estimation, facilitating the computation of cosine similarities between test images and class distributions. Even though we lower the bias between the estimated distributions and the real distributions by augmenting the support set, the amount of samples used to estimate the distributions remains limited, which means the bias still exists, To further push the estimated distribution closer to the real distribution. Given the image feature $f_v$ extracted by $E_{v}$, the logits of visual adapter can be denoted as
\begin{equation}
\label{eq-Lv}
    L_v = Q\left( f_v \right) =\exp \left( -\eta (1-Wf_v) \right) .
\end{equation}
During training, $W$ is updated by gradient descent.

\textbf{Textual Adapter.}
In TT-DNA, we design the prompt and class name manually. CoOp \cite{zhou2022learning} has demonstrated that the prompt learning method surpasses the performance of manually created prompts. Therefore, we can also improve the model's performance by prompt tuning. 
However, CoOp employs a pre-model prompting technique, which during training, may be influenced by prior knowledge, as seen in CoOp's 1-shot classification accuracy being lower than that of Zero-shot CLIP. We adopt a post-model prompting strategy by adding a textual adapter which is a learnable matrix to the text features. During training, we only tune the textual adapter while keeping the text features fixed, enabling reliable original CLIP knowledge preservation and learning an optimal text feature. As illustrated in Figure \ref{fig:overview}, we add a learnable matrix (\textit{i.e.} the textual adapter) to the text features $f_t$ generated by text encoder $E_t$. The new text features is denoted as $\hat{f}_t=f_t+\beta A$.
$A$ is a learnable matrix, $\beta$ is a hyper-parameter that controls the scaling of residual. The logit of enhanced CLIP is:
\begin{equation}
\label{eq-LC}
    L_C=f_v{\hat{f}_t}^{\top}=f_v(f_t+\beta A)^{\top}.
\end{equation} 
Finally, the total logits of the input image $v$ by TT-DNA-F are 
\begin{align}
\label{eq-Lc}
\mathrm{Logits} &= \lambda L_v+L_C \nonumber \\ 
                &=\lambda \exp \left( -\eta \left( 1-Wf_v \right) \right) +f_v(f_t+\beta A)^{\top}.
\end{align}
where $\lambda$ is a hyper-parameter to control the scaling of the residual, $W$ and $A$ are learnable parameters. 

\section{Experiment}

\begin{table}[t]
\vspace{0.7pt}
\caption{Performance comparisons of our TT-DNA methods and other baselines on the Bongard-HOI \cite{jiang2022bongard} dataset.}
\vspace{1pt}
\label{table:hoi}
\centering
\resizebox{\linewidth}{!}{
\begin{tabular}{lccccc}
\toprule
\multirow{3}{*}{Method} & \multicolumn{5}{c}{Test Splits}          \\
\cmidrule(lr){2-6}
                        & Seen act.     & Unseen act.     & Seen act.     & Unseen act.     & \multirow{2}{*}{Avg.}    \\
                        & Seen obj.     & Seen obj.     & Unseen obj.     & Unseen obj.     &     \\
\midrule
CNN-Baseline \cite{nie2020bongard}                   & 50.03 & 49.89 & 49.77 & 50.01 & 49.92 \\
Meta-Baseline \cite{chen2020new}                   & 58.82 & 58.75 & 58.56 & 57.04 & 58.30 \\
ProtoNet  \cite{snell2017prototypical}                  & 58.90 & 58.77 & 57.11 & 58.34 & 58.28 \\
HOITrans \cite{zou2021end}                    & 59.50 & 64.38 & 63.10 & 62.87 & 62.46 \\
TPT  \cite{shu2022tpt}                    & 66.39 & 68.50 & 65.98 & 65.48 & 66.59 \\
BDC-Adapter \cite{zhang2023bdc}                    & 68.36 & 69.15 & 67.67 & 67.82 & 68.25 \\
\rowcolor{gray!20}
TT-DNA (Ours)    & \textbf{68.85} & \textbf{70.16} & \textbf{68.97} & \textbf{67.33} & \textbf{68.83} \\
\rowcolor{gray!20}
TT-DNA-F (Ours)    & \textbf{69.45} & \textbf{71.56} & \textbf{70.07} & \textbf{69.76} & \textbf{70.21} \\
\bottomrule
\end{tabular}
}
\vspace{-12pt}
\end{table}

\subsection{Visual Reasoning on Bongard-HOI} \label{sec:cdvs}
\textbf{Baselines.} 
We consider five previous methods for comparison: 
(1) The CNN-Baseline \cite{nie2020bongard} is a basic classifier trained on Bongard-HOI data. It's trained to map complete training samples into a binary output.
(2) The Meta-Baseline \cite{chen2020new} treats individual samples in Bongard-HOI as few-shot tasks and endeavors to rapidly adapt the model to novel tasks.
(3) ProtoNet \cite{snell2017prototypical} facilitates classification by calculating distances from prototype representations for every class.
(4) 
HOITrans \cite{zou2021end} is a transformer-driven model for HOI detection, resolving Bongard-HOI by contrasting the detected HOIs in query images with those in the support images.
(5) 
TPT \cite{shu2022tpt} is built upon CLIP and has the capability to adaptively learn prompts dynamically using a single test sample.
(6) BDC-Adapter \cite{zhang2023bdc}, which introduces Brownian Distance Covariance to visual reasoning task.


\textbf{Implementation Details.} 
Our approach is based on the CLIP model, with ResNet-50 as its image encoder and transformer as its text encoder. Both visual and text encoders of CLIP are frozen during fine-tuning. We follow the data pre-processing protocol in CLIP.  
We initialize the learnable matrix with zeros. In Equation (\ref{eq-LC}), we set $\beta$ to 0.6. We use an initial learning rate of $10^{-3}$, and optimize our models with the AdamW optimizer. 
It is different between visual reasoning on HOI and traditional image classification, The correctness of predictions in Bongard-HOI relies on binary contexts, indicating the presence or absence of a concept (c). When dealing with binary labels, a direct approach is to manually create "labels" for positive and negative examples. Here, we use \texttt{True} or \texttt{False} as labels. We construct a hand-crafted prompt $\rho =$ ``a photo that a person \{\texttt{action}\} \{\texttt{object}\}, it is \{\texttt{class}\}", where \{\texttt{action}\} is the action  such as ``eat", ``squeeze", ``sit on", \textit{etc.}; \{\texttt{object}\} is the object in the image such as ``orange", ``bicycle", \textit{etc.}; \{\texttt{class}\} is ``true" or ``false". Here we give an example of the prompt: ``a photo that a person eats orange, it is true."

\textbf{Results.} 
We compare the performance of Our proposed method with previous methodologies, as presented in Table \ref{table:hoi}. Remarkably, our proposed TT-DNA outperforms all the conventional non-CLIP methods by large margins. Even compared to the CLIP-based TPT method, the training-free TT-DNA still yields superior performance than TPT. In comparison with BDC-Adapter, TT-DNA performs much better in three test scenarios but slightly worse in one test scenario. With training, TT-DNA-F achieves remarkable performance gain, outperforming TPT and BDC-Adapter by large margins in all the test scenarios. It demonstrates the effectiveness of the query model and textual adapter. In detail, the query model can further push the estimated class distribution closer to the expected class distribution, and the textual adapter can learn optimal text features.
In accordance with the experimental design outlined in Jiang \textit{et al.} \cite{jiang2022bongard}, 
The evaluation encompasses four separate test splits within the Bongard-HOI dataset. It's worth noting that the outcomes of the other baseline methods are directly extracted from Jiang \textit{et al.} \cite{jiang2022bongard}. 

\subsection{Ablation Study}
\textbf{Effectiveness of Different Algorithm Components.}
Our method is built upon CLIP,  we compare its different components appended to CLIP with four test splits. As shown in Table \ref{table:component}, we find that both the query model and textual adapter bring performance gain over standalone CLIP.  This ablation study suggests that both components contribute significantly to the overall performance. 

\textbf{Visual Backbone.}
Table \ref{table:backbones} summarizes the average results on four test splits using various visual backbones containing ResNets and ViTs. We can see that our method yields better performance on more advanced visual backbones.

\begin{table}[t]
\caption{Effectiveness of different algorithm components in our TT-DNA-F method. In this table, QM represents the query model, and TA represents the textual adapter.}
\vspace{1pt}
\label{table:component}
\centering
\resizebox{\linewidth}{!}{
\begin{tabular}{lccccc}
\toprule
\multirow{3}{*}{Method} & \multicolumn{5}{c}{Test Splits}          \\
\cmidrule(lr){2-6}
                        & Seen act.     & Unseen act.     & Seen act.     & Unseen act.     & \multirow{2}{*}{Avg.}    \\
                        & Seen obj.     & Seen obj.     & Unseen obj.     & Unseen obj.     &     \\
\midrule
Zero-shot CLIP \cite{radford2021learning}                   & 63.38 & 65.29 & 63.77 & 64.05 & 64.12 \\
\quad  + TA                   & 66.95 & 69.38 & 67.51 & 68.05 & 67.97 \\
\quad  + QM                  & 69.07 & 70.68 & 69.28 & 68.25 & 69.32 \\
\rowcolor{gray!20}
\quad  + TA + QM (Ours)    & \textbf{69.45} & \textbf{71.56} & \textbf{70.07} & \textbf{69.76} & \textbf{70.21} \\
\bottomrule
\end{tabular}
}
\vspace{-6pt}
\end{table}

\begin{table}[t]
\caption{Evaluation of our method on various visual backbones on four test splits. We use the average accuracy for comparison.}
\vspace{1pt}
\label{table:backbones}
\centering
\resizebox{\linewidth}{!}{
\begin{tabular}{lcccc}
\toprule
\multirow{2}{*}{Method} & \multicolumn{4}{c}{Visual Backbone}          \\
\cmidrule(lr){2-5}
                        & ResNet-50 & ResNet-101 & ViT-B/32 & ViT-B/16 \\
\midrule
TT-DNA           & 68.83     & 69.92      & 70.15    & 71.93    \\
\rowcolor{gray!20}
TT-DNA-F                   & \textbf{70.21}     & \textbf{71.87}      & \textbf{72.15}    &\textbf{74.52}     \\
\bottomrule
\end{tabular}
}
\vspace{-8pt}
\end{table}

\section{Conclusion}
We present a novel method called Test-time Distribution Learning Adapter (TT-DNA) to further explore the potential of CLIP. We use Gaussian distributions to model visual features of few-shot support images, capturing support set knowledge. The visual adapter predicts using cosine similarity between the query image and support image features. It's then combined with CLIP's prediction via a residual connection, yielding the final prediction. Extensive experiments show TT-DNA outperforms existing methods by a significant margin in human-object interaction visual reasoning.


\clearpage
\bibliographystyle{IEEEbib}
\bibliography{strings,refs}

\begin{thebibliography}{10}

\bibitem{radford2021learning}
Alec Radford, Jong~Wook Kim, Chris Hallacy, Aditya Ramesh, Gabriel Goh, Sandhini Agarwal, Girish Sastry, Amanda Askell, Pamela Mishkin, Jack Clark, et~al.,
\newblock ``Learning transferable visual models from natural language supervision,''
\newblock in {\em International conference on machine learning}. PMLR, 2021, pp. 8748--8763.

\bibitem{jia2021scaling}
Chao Jia, Yinfei Yang, Ye~Xia, Yi-Ting Chen, Zarana Parekh, Hieu Pham, Quoc Le, Yun-Hsuan Sung, Zhen Li, and Tom Duerig,
\newblock ``Scaling up visual and vision-language representation learning with noisy text supervision,''
\newblock in {\em International Conference on Machine Learning}, 2021, pp. 4904--4916.

\bibitem{gao2021clip}
Peng Gao, Shijie Geng, Renrui Zhang, Teli Ma, Rongyao Fang, Yongfeng Zhang, Hongsheng Li, and Yu~Qiao,
\newblock ``Clip-adapter: Better vision-language models with feature adapters,''
\newblock {\em International Journal of Computer Vision}, pp. 1--15, 2023.

\bibitem{shi2022proposalclip}
Hengcan Shi, Munawar Hayat, Yicheng Wu, and Jianfei Cai,
\newblock ``Proposalclip: Unsupervised open-category object proposal generation via exploiting clip cues,''
\newblock in {\em Proceedings of the IEEE/CVF Conference on Computer Vision and Pattern Recognition}, 2022, pp. 9611--9620.

\bibitem{du2022learning}
Yu~Du, Fangyun Wei, Zihe Zhang, Miaojing Shi, Yue Gao, and Guoqi Li,
\newblock ``Learning to prompt for open-vocabulary object detection with vision-language model,''
\newblock in {\em Proceedings of the IEEE/CVF Conference on Computer Vision and Pattern Recognition}, 2022, pp. 14084--14093.

\bibitem{yao2021cpt}
Yuan Yao, Ao~Zhang, Zhengyan Zhang, Zhiyuan Liu, Tat-Seng Chua, and Maosong Sun,
\newblock ``Cpt: Colorful prompt tuning for pre-trained vision-language models,''
\newblock {\em arXiv preprint arXiv:2109.11797}, 2021.

\bibitem{zhou2022learning}
Kaiyang Zhou, Jingkang Yang, Chen~Change Loy, and Ziwei Liu,
\newblock ``Learning to prompt for vision-language models,''
\newblock {\em International Journal of Computer Vision}, vol. 130, no. 9, pp. 2337--2348, 2022.

\bibitem{lu2022prompt}
Yuning Lu, Jianzhuang Liu, Yonggang Zhang, Yajing Liu, and Xinmei Tian,
\newblock ``Prompt distribution learning,''
\newblock in {\em Proceedings of the IEEE/CVF Conference on Computer Vision and Pattern Recognition}, 2022, pp. 5206--5215.

\bibitem{chen2023plot}
Guangyi Chen, Weiran Yao, Xiangchen Song, Xinyue Li, Yongming Rao, and Kun Zhang,
\newblock ``{PLOT}: Prompt learning with optimal transport for vision-language models,''
\newblock in {\em International Conference on Learning Representations}, 2023.

\bibitem{zhang2022tip}
Renrui Zhang, Wei Zhang, Rongyao Fang, Peng Gao, Kunchang Li, Jifeng Dai, Yu~Qiao, and Hongsheng Li,
\newblock ``Tip-adapter: Training-free adaption of clip for few-shot classification,''
\newblock in {\em European Conference on Computer Vision}. Springer, 2022, pp. 493--510.

\bibitem{jiang2022bongard}
Huaizu Jiang, Xiaojian Ma, Weili Nie, Zhiding Yu, Yuke Zhu, and Anima Anandkumar,
\newblock ``Bongard-hoi: Benchmarking few-shot visual reasoning for human-object interactions,''
\newblock in {\em Proceedings of the IEEE/CVF Conference on Computer Vision and Pattern Recognition}, 2022, pp. 19056--19065.

\bibitem{shu2022tpt}
Shu Manli, Nie Weili, Huang De-An, Yu~Zhiding, Goldstein Tom, Anandkumar Anima, and Xiao Chaowei,
\newblock ``Test-time prompt tuning for zero-shot generalization in vision-language models,''
\newblock in {\em Advances in Neural Information Processing Systems}, 2022, pp. 14274--14289.

\bibitem{zhang2023bdc}
Yi~Zhang, Ce~Zhang, Zihan Liao, Yushun Tang, and Zhihai He,
\newblock ``Bdc-adapter: Brownian distance covariance for better vision-language reasoning,''
\newblock in {\em British Machine Vision Conference}, 2023.

\bibitem{nie2020bongard}
Weili Nie, Zhiding Yu, Lei Mao, Ankit~B Patel, Yuke Zhu, and Anima Anandkumar,
\newblock ``Bongard-logo: A new benchmark for human-level concept learning and reasoning,''
\newblock in {\em Advances in Neural Information Processing Systems}, 2020, vol.~33, pp. 16468--16480.

\bibitem{chen2020new}
Yinbo Chen, Zhuang Liu, Huijuan Xu, Trevor Darrell, and Xiaolong Wang,
\newblock ``Meta-baseline: Exploring simple meta-learning for few-shot learning,''
\newblock in {\em Proceedings of the IEEE/CVF International Conference on Computer Vision}, 2021, pp. 9062--9071.

\bibitem{liu2021domain}
Yajing Liu, Zhiwei Xiong, Ya~Li, Xinmei Tian, and Zheng-Jun Zha,
\newblock ``Domain generalization via encoding and resampling in a unified latent space,''
\newblock {\em IEEE Transactions on Multimedia}, vol. 25, pp. 126--139, 2021.

\bibitem{liu2022category}
Yajing Liu, Zhiwei Xiong, Ya~Li, Yuning Lu, Xinmei Tian, and Zheng-Jun Zha,
\newblock ``Category-stitch learning for union domain generalization,''
\newblock {\em ACM Transactions on Multimedia Computing, Communications and Applications}, vol. 19, no. 1, pp. 1--19, 2023.

\bibitem{zhang2021adversarial}
Yonggang Zhang, Mingming Gong, Tongliang Liu, Gang Niu, Xinmei Tian, Bo~Han, Bernhard Sch{\"o}lkopf, and Kun Zhang,
\newblock ``Adversarial robustness through the lens of causality,''
\newblock in {\em International Conference on Learning Representations}, 2021.

\bibitem{wang2021regularizing}
Yulin Wang, Gao Huang, Shiji Song, Xuran Pan, Yitong Xia, and Cheng Wu,
\newblock ``Regularizing deep networks with semantic data augmentation,''
\newblock {\em IEEE Transactions on Pattern Analysis and Machine Intelligence}, vol. 44, no. 7, pp. 3733--3748, 2021.

\bibitem{zhang2023prompt}
Renrui Zhang, Xiangfei Hu, Bohao Li, Siyuan Huang, Hanqiu Deng, Yu~Qiao, Peng Gao, and Hongsheng Li,
\newblock ``Prompt, generate, then cache: Cascade of foundation models makes strong few-shot learners,''
\newblock in {\em Proceedings of the IEEE/CVF Conference on Computer Vision and Pattern Recognition}, 2023, pp. 15211--15222.

\bibitem{Mini_Dalle}
Boris Dayma, Suraj Patil, Pedro Cuenca, Khalid Saifullah, Tanishq Abraham, Phuc Le~Khac, Luke Melas, and Ritobrata Ghosh,
\newblock ``Dall·e mini,'' 2021.

\bibitem{ru2022learning}
Lixiang Ru, Yibing Zhan, Baosheng Yu, and Bo~Du,
\newblock ``Learning affinity from attention: End-to-end weakly-supervised semantic segmentation with transformers,''
\newblock in {\em Proceedings of the IEEE/CVF Conference on Computer Vision and Pattern Recognition}, 2022, pp. 16846--16855.

\bibitem{snell2017prototypical}
Jake Snell, Kevin Swersky, and Richard Zemel,
\newblock ``Prototypical networks for few-shot learning,''
\newblock in {\em Advances in Neural Information Processing Systems}, 2017, vol.~30, pp. 4080--4090.

\bibitem{zou2021end}
Cheng Zou, Bohan Wang, Yue Hu, Junqi Liu, Qian Wu, Yu~Zhao, Boxun Li, Chenguang Zhang, Chi Zhang, Yichen Wei, et~al.,
\newblock ``End-to-end human object interaction detection with hoi transformer,''
\newblock in {\em Proceedings of the IEEE/CVF Conference on Computer Vision and Pattern Recognition}, 2021, pp. 11825--11834.

\end{thebibliography}

\end{document}